\documentclass{article}

\usepackage{arxiv}
\usepackage[utf8]{inputenc} 
\usepackage[T1]{fontenc}    
\usepackage{hyperref}       
\usepackage{url}            
\usepackage{booktabs}       
\usepackage{amsfonts}       
\usepackage{nicefrac}       
\usepackage{microtype}      
\usepackage{natbib}
\usepackage{doi}

\usepackage{amsmath}
\usepackage{cleveref}
\usepackage[font=it]{caption} 
\usepackage{multicol,lipsum,graphicx,float} 

\begin{document}
\title{Guided Transfer Learning}
\author{ 
\hspace{1mm}Danko Nikolić
\thanks{Sachsenhäuser Landwehrweg 96, 60599 Frankfurt am Main, Germany (www.robotsgomental.com)} 
\\
	Robots Go Mental\\
	\texttt{danko@robotsgomental.com} \\
	 \And Davor Andrić \\
	Robots Go Mental\\
	\texttt{davor@robotsgomental.com} \\
	 \And Vjekoslav Nikolić\\
	Robots Go Mental \\
	\texttt{vjeko@robotsgomental.com} \\
}
\maketitle

\renewcommand{\shorttitle}{Guided transfer learning}

\hypersetup{
pdftitle={Guided Transfer Learning},
pdfsubject={},
pdfauthor={},
pdfkeywords={},
}

\begin{abstract}
Machine learning requires exuberant amounts of data and computation. Also, models require equally excessive growth in the number of parameters. It is, therefore, sensible to look for technologies that reduce these demands on resources. Here, we propose an approach called guided transfer learning. Each weight and bias in the network has its own \emph{guiding} parameter that indicates how much this parameter is allowed to change while learning a new task. Guiding parameters are learned during an initial scouting process. Guided transfer learning can result in a reduction in resources needed to train a network. In some applications, guided transfer learning enables the network to learn from a small amount of data. In other cases, a network with a smaller number of parameters can learn a task which otherwise only a larger network could  learn. Guided transfer learning potentially has many applications when the amount of data, model size, or the availability of computational resources reach their limits.
\end{abstract}


\begin{multicols}{2}
\section{Introduction}
The problem of learning in artificial neural networks requires finding a specific set of values for connection weights and biases such that the network as a whole accurately performs a required task [1]. Finding the right set of values for these parameters is often not an easy task, especially when the networks become large and the tasks become ambitious. The main reason for the difficulties in learning is that the space of possibilities, the so-called parameter space, tends to be huge, especially in deep learning, which may have millions or even billions of parameters. This means that the entire space can never be explored to find the most optimal set of parameters. Rather, only a small fraction of possible values can be evaluated. Hence, the learning algorithms rely on making small, informed steps toward a possible solution. The algorithms search for ‘hints’ on the direction in which the values in the parameters space should be changed. Then this new state is evaluated, and new hints are calculated. And so on. These hints are calculated based on the training data. At each iteration, a question is asked: What do training data suggest? In which direction should we move the weights and biases to find a solution? This learning method is known as gradient descent and is the most popular approach for training deep learning networks.
One of the issues that have recently become obvious is that the deep learning models present unsustainable demands on resources. The most advanced deep learning models require too large a number of parameters and large training data sets to make such neural networks affordable for everyone. Instead, only well-endowed institutions can afford to develop such models. This problem is well substantiated by a study by Kaplan et al. [2] on how the demands on language models grow. They showed that an increase in the amount of data helps only until a point, after which it does not help any more unless the model size is increased (i.e., the number of parameters is increased); thereafter, subsequent increases in model sizes keep improving the performance but only up to a point where again an increase in the data size is needed to accompany the increase in the model size. And so on (see Figure 1A). Apparently, there is no limit in intelligence that these models can reach, provided that we continue increasing both the amount of data and the model sizes. A later study reported that one can trade off data size and model size to a degree: A somewhat smaller model can be trained by disproportionally increasing the amount of data [3].
Unfortunately, however, a conclusion that follows from Kaplan et al.’s [2] study, but which was not addressed in their publication, is that the price paid for this apparently unlimited increase in intelligence increase is so high that the entire endeavor becomes unsustainable: Very soon, for a miniature increase in intelligence, a correspondingly large amount of resources needs to be provided. This is because, as Kaplan et al. found, demands on resources grow with a power law. This means that the sizes of the models explode and so do the amounts of data required: To double the intelligence of a model, i.e. to reduce the loss by half, one needs a lot more than double the amount of resources. The large exponent of a power law $\gg{1}$ is why language models today are so large, and only institutions with deep pockets can afford to build them. Also, this is why these models cannot grow in intelligence much further; there are not enough resources in the world. A study in computer vision found that the situation is not any better there either. Models that recognize objects from images require about a 500-fold increase in resources to only double their intelligence [4]. Other studies reached similar conclusions [5,6] (see [7] for a review of how machine learning engineers and data scientists cope in practice with these limitations of deep learning models).

\subsection{Background}
\label{sec:headings}
There is a theoretical possibility that much smaller deep learning models exist and are able to perform exactly the same tasks as the large ones – the only problem being the inability of the learning algorithms to find the parameter values for such smaller models. One known example is the n-parity problem, also known as the generalized XOR problem. This can be implemented by small neural networks, however, these small networks have to be either constructed by hand [8] or learned by specialized learning algorithms designed specifically for that problem [9,10,11]. These specialized learning algorithms do not rely on gradient descent and hence, can learn only a handful of other functions besides that one. In contrast, gradient descent algorithms are able to train neural networks to perform a variety of tasks, including many real-life tasks. Gradient descent can also learn to solve n-parity problems but the amount of required resources explodes. A problem that can be solved by only a handful of parameters when constructed by hand may require 100s or even millions of parameters in deep learning networks trained through gradient descent [8].

\begin{figure}[H]
   \includegraphics[width=8.5cm]{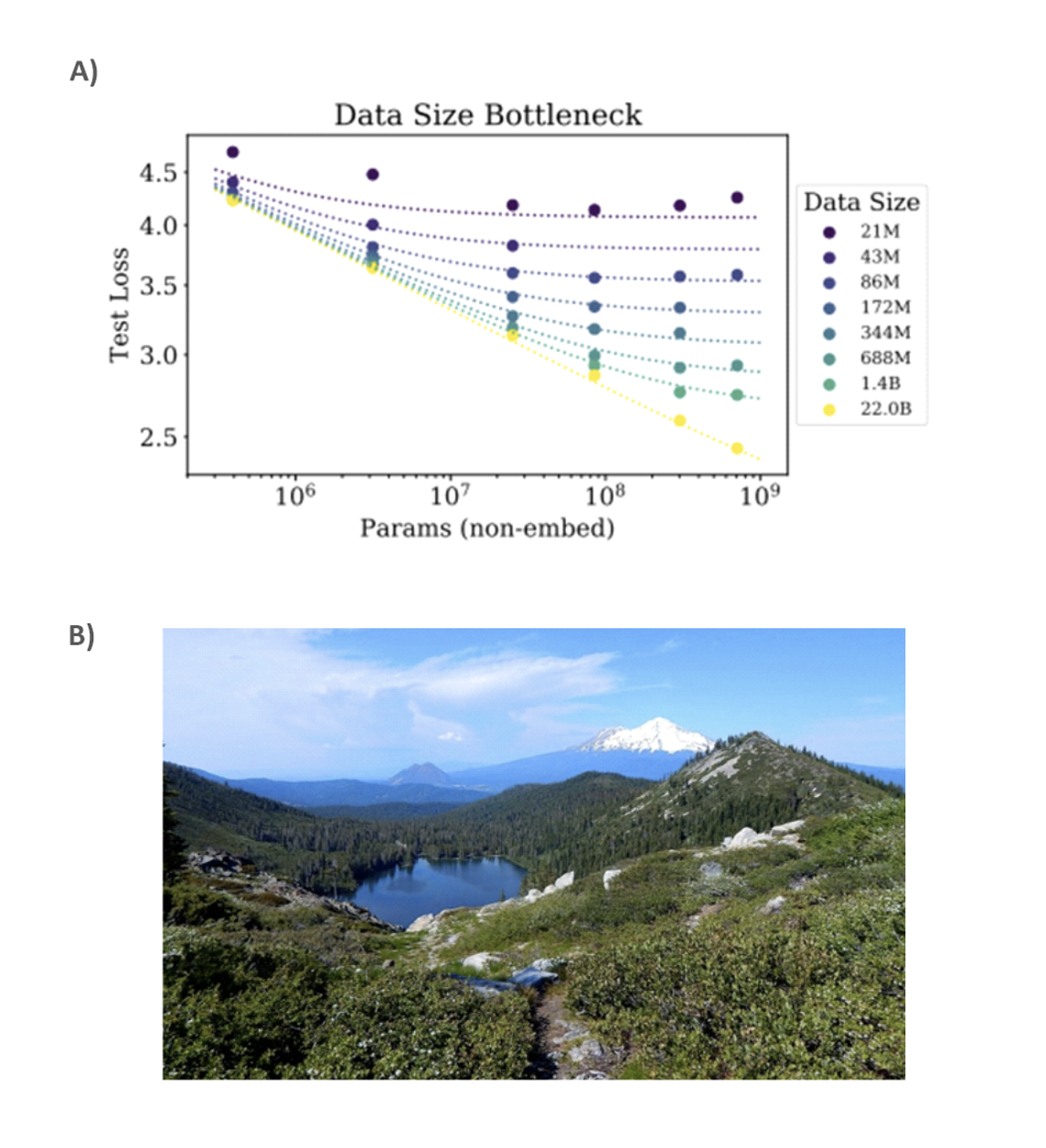}
   \caption{Figure 1. Sizes of deep learning models explode and so do the demands for data. A) The performance of large language models (Test Loss) depends on the resources: the number of parameters in a model and the amount of data. Note the logarithmic scales of all variables, indicating that demands on resources explode. A straight line in a log-log plot (the yellow line) indicates a power law relationship. Adapted from [2]. B) Local minima forming during training of deep learning networks are like mountain lakes. The rainwater descends down a mountain by only looking locally. Consequently, if there is a local minimum, the water gets stuck. Deep Learning traditionally solves this local minimum problem by adding more dimensions (water has only two dimensions along which to move. Deep learning can have millions or billions of dimensions, and these additional dimensions make local minima (i.e., mountain lakes) less likely to form. This necessarily requires a corresponding increase in the amount of data. Guided transfer learning avoids local minima by sending scouts to look behind the lakes and then telling the main model which dimensions to ignore, i.e. to descend over only a subset of the dimensions. This is like walking along a slope of a mountain rather than walking straight down to the nearest valley. In some cases, the valley is a trap, a local minimum trap, and scouts know that. Image source: https://rootsrated.com/redding-ca/hiking/castle-lake-to-heart-lake-trail-hiking (labeled as creative common license).}
   \label{fig:long1}
\end{figure}

Gradient descent simply has too short a “depth of view”; while finding hints on how to change parameters, the algorithm cannot see how changes now will affect the performance later down the descent path. In other words, gradient descent cannot know whether the choices made now will make it end up in a local minimum later (Figure 1B). We think that the way gradient descent works around these problems is highly inefficient: It increases the number of parameters together with the amount of data. By a sheer increase in the number of dimensions it becomes less likely that the model will end up in a local minimum. As more parameters make it more likely to overfit by poorly generalizing i.e., by choosing too easy paths downhill during the descent, the increase in parameters needs to be matched with a corresponding increase in the amount of training data. Therefore, the lack of limits in intelligence that deep learning can achieve is overcome by a two-step brute force approach: increase dimensions to avoid local minima, and increase the amount of data to avoid overfitting. Unfortunately, however, the prices paid for this unlimited intelligence are too high to be sustainable, resulting in exploding model sizes and a hunger for data.

Clearly, there is a need to assist gradient descent with a better look into what lies ahead of the effects of the current parameter changes on the model. Any informed help for ‘looking more deeply’, i.e., down the path of descent, may considerably reduce the number of parameters and data points required for successful training. A related example is the addition of inertia to learning mechanisms, which is another method to avoid local minima. The idea is that if you encounter a local minimum you just run over it by keeping the same direction you have been traversing so far [12]. Inertia algorithms hope to avoid local minima by allowing for no sharp turns. However, inertia only refers to information from the values of the parameters at already visited locations; it does not know anything about what is to be expected ahead.

The performance of large language models (Test Loss) depends on the resources Figure 1, A) the number of parameters in a model and the amount of data. Note the logarithmic scales of all variables, indicating that demands on resources explode. A straight line in a log-log plot (the yellow line) indicates a power law relationship. Adapted from [2]. B) Local minima forming during training of deep learning networks are like mountain lakes. The rainwater descends down a mountain by only looking locally. Consequently, if there is a local minimum, the water gets stuck. Deep Learning traditionally solves this local minimum problem by adding more dimensions (water has only two dimensions along which to move. Deep learning can have millions or billions of dimensions, and these additional dimensions make local minima (i.e., mountain lakes) less likely to form. This necessarily requires a corresponding increase in the amount of data. Guided transfer learning avoids local minima by sending scouts to look behind the lakes and then telling the main model which dimensions to ignore, i.e. to descend over only a subset of the dimensions. This is like walking along a slope of a mountain rather than walking straight down to the nearest valley. In some cases, the valley is a trap, a local minimum trap, and scouts know that. Image source: https://rootsrated.com/redding-ca/hiking/castle-lake-to-heart-lake-trail-hiking (labeled as creative common license).
 
\section{Guidance matrix}
The guidance matrix results from the scouting process and guides the model during training. The guidance matrix works simply by providing a number informing the model how important a certain dimension is, i.e., a certain parameter. If the value is low, say, close to zero, scouts did not find many changes along this dimension (parameter). Therefore, the main model may also want to ignore this parameter even if its data locally suggest that changes should be made. In effect, the scouts are saying: “We have tried this parameter, but it did not work for us. It is better not to make any changes here.” In contrast, if the value in the guidance matrix is large, the scouts are saying: “We found that it is a good idea to change this parameter.”

Mathematically, guidance is a simple operation. Let us say that the guiding value for a weight \(w\) is \(g_w\). Then, if gradient descent computes a value of changing \(\Delta w\), the guided change \(\Delta w_g\) will be computed as:

\begin{equation*}
\Delta w_{g}=\Delta w\  \times g_{w}
\end{equation*}
The same holds for changes in a bias:
\begin{equation*}
\Delta b_{g}=\Delta b\  \times g_{b}
\end{equation*}
That is all. This operation can be expressed in the matrix calculus as:
\begin{equation*}
\Delta W_{g}=\Delta W\  \times G_{W}
\end{equation*}
where \(G_w\) is the guidance matrix. Application of the guidance matrix is also easy to implement into a deep learning code (Figure 2).

\begin{figure}[H]
\includegraphics[width=8cm]{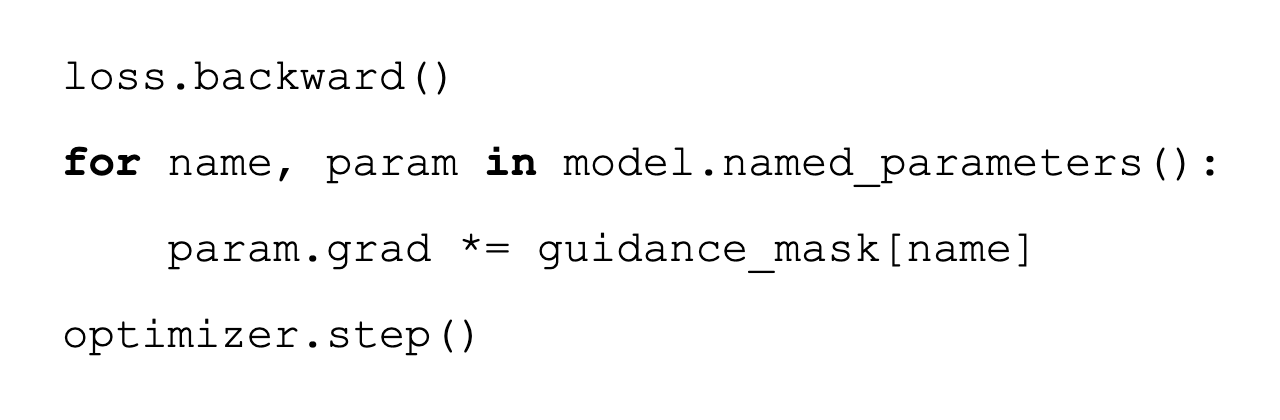}
   \caption{In PyTorch, the application of the guiding matrix can be implemented by adding two lines of code between the backward computation of all deltas (loss.backward()) and the application of those deltas (optimizer.step()).}
   \label{fig:long2}
   \label{fig:onecol}
\end{figure}

The guidance matrix contains exactly one guidance parameter for each model parameter that is being learned. Therefore, guided transfer learning requires as much additional memory as it takes to store the main model. 

The possibilities of guided transfer learning are not exhausted by providing a single guidance matrix. More elaborated forms of guidance matrices are possible. For example, scouts may provide different guidance matrices at different depths i.e., at different distances from the starting point. Possibilities are diverse, and only limited by our creativity.

\subsection{Guided transfer learning complements transfer learning by pre-training}

Classically, transfer learning is made by pre-training a model on another related data set before training is continued with the main task. That way, transfer learning by pre-training helps avoid local minima simply by bringing the state of the network into a relative vicinity of the solution that the main problem will require. Through pre-training, many of the local minima have already been dealt with. The data set for pre-training is often much larger than that of the later task. Consequently, pre-training effectively avoids a great proportion of all the local minima lurking during the training of the main task. 

Unfortunately, transfer learning by pre-training does not help with the remaining local minima that will be encountered during the subsequent training on the main task. This is where Guided transfer learning helps. Consequently, the two forms of transfer learning–pre-training and guided learning–can be combined (but do not always have to be), complementing each other. One provides a great starting point for learning, and the other guides the directions of this continued learning. 
. 

\subsection{Scouting} 
Scouting is a process necessary for the creation of guidance matrices. Scouting looks further into the space of parameters in order to explore “what lies behind them”. In a way, scouting looks at what is expected to be encountered in the ‘future’. 
Scouting is performed by replacing a difficult problem with multiple easier ones. One way to make scouting easier is to make the task less difficult to solve. For example, the task of a scout may require considerably fewer categories to learn. An additional way to make scouting easier to learn is to use a training data set with far more data points. The latter is similar to transfer learning by pre-training: Fewer categories and more data points reduce the chances of hitting a local minimum. 

So, for example, a 10-category problem can be broken into multiple 3-category problems, each problem achieving better accuracy than the big problem with 10 categories. Scouts are allowed to share categories. When scouts share categories, they are related and we refer to them as cousins – we form an entire family of small problems (Figure 3A). 

All scouts start learning from the same base model with parameters \(w_B\) and \(b_B\). From that point on, new parameters are learned after some learning criterion. Let us denote those as \(w_1 ... w_n\), and \(b_1 ... b_n\) where \(n\) is the number of scouts (Figure 3B). 

We then compute the guidance matrix, \(G\). There are multiple ways to compute \(G\), each with advantages and disadvantages. The simplest way to obtain the guidance values is to calculate the average squared distance from the starting point. We first obtain the mean deviation for each weight, \(m\): 
\begin{equation*}
m_{w} = \frac {1}{n} \Sigma ( w_{B} - w_{i})^2
\end{equation*}
In matrix form, mean deviations for weights and biases are denoted as \(M_w\) and \(M_b\), respectively.
We finally need to normalize M to obtain G. The normalization choice will affect some properties of the guidance matrix such as the learning rate. A slower, conservative learning rate will be obtained if normalizing by a scalar that represents the maximum value in M: 
\begin{equation*}
G_{w} = \frac {M_w}{max(M_w)}; G_{b} = \frac {M_b}{max(M_b)}
\end{equation*}
Faster learning will be achieved by normalizing by some form of indicator of a mean value, for example:
\begin{equation*}
G = \frac {M}{mean(M)}
\end{equation*}
We can also apply a forced-zero approach to normalization in which the guidance parameters contain at least one value of 0. For example, the smallest value can be forced to 0 and the largest to 1, the rest being accordingly scaled between those two values: 
\begin{equation*}
G = \frac {M-min(M)}{max(M) - min(M)} 
\end{equation*}
An example of a resulting distribution of the values in \(G\) is given in Figure 3C.
Squaring the difference between \(w_B\) and \(w_i\) emphasizes big changes over small changes. To treat all the differences more equally it is possible to compute \(M\) by summing up absolute differences instead of squared differences: 
\begin{equation*}
m_{W} = \frac {1}{n} \Sigma \vert w_{s} - w_{i}\vert
\end{equation*}
As mentioned above, more elaborate forms of \(G\) are possible for example, by computing a separate \(G\) for each training epoch.
\begin{figure}[H]
\begin{center}
   \includegraphics[width=7.5cm]{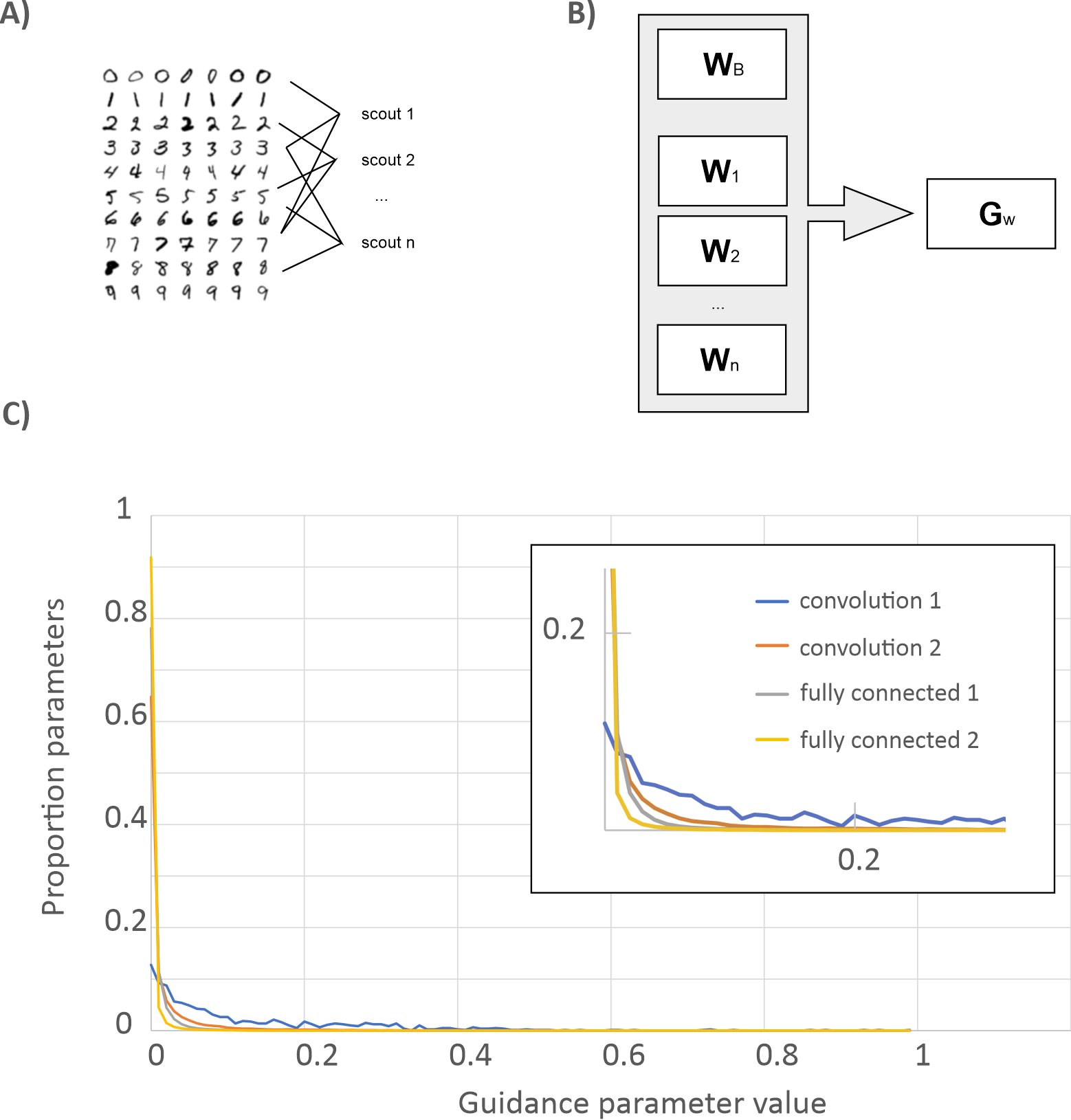}
\end{center}
   \caption{Calculation of a guidance matrix. A) Example formation of cousin scouts from the MNIST data set. B) The Guidance matrix is computed from the base matrix, \(W_B\), and the scout matrices \(W_1 ... W_n\). C) Typical distributions of guidance values. Inset: A zoom-in with more detail. Majority of parameters have very small guiding values. Only a few parameters have large values.}
\label{fig:long3}
\end{figure}

The distributions of obtained \(g\) values across the entire \(G\) tend to be quite narrow: A small number of values are high, and a large number of guidance values \(g\) are low (as in Figure 3C). Such a distribution indicates that matrix \(G\) will strongly affect training. When this is the case, effectively, the model becomes a sparse model during guided training, with only a few parameters allowed to change. Thus, the learning problem is reduced to a smaller number of dimensions. The computations of the network remain dense; it is only the training that becomes sparse.
Example code implementing guided transfer learning can be found here: \url{https://github.com/RobotsGoMental/gtl_poc}

\subsection{Example applications}

\subsubsection{Avoiding overfit in a one-shot learning task}
Often there is insufficient data available for training a model. The most extreme case is the so-called one-example learning or “one-shot learning”. In that case, a single training example would be given per category and the model would be then tested on multiple examples to assess the performance accuracy. A data set commonly used for such a task is Omniglot [15,16], which contains a variety of alphabets (Latin, Cyrillic, Glagolitic, and many more; in total 50) and 20 hand-written examples of each character in each of the alphabets. The task is then to train a model on a single example and test the ability to recognize other hand-written versions of this letter among other letters of the same alphabet (Figure 4A).

We used a combination of transfer learning by pre-training and guided transfer learning. To pre-train the model and to compute the guiding matrices for the model parameters, we used another dataset containing human handwriting MNIST, which contains only 10 characters (digits 0 to 9) but has a large number of training examples per character, namely 10,000 [17]. It is easy to obtain more than 90

\begin{figure}[H]
   \includegraphics[width=7.5cm]{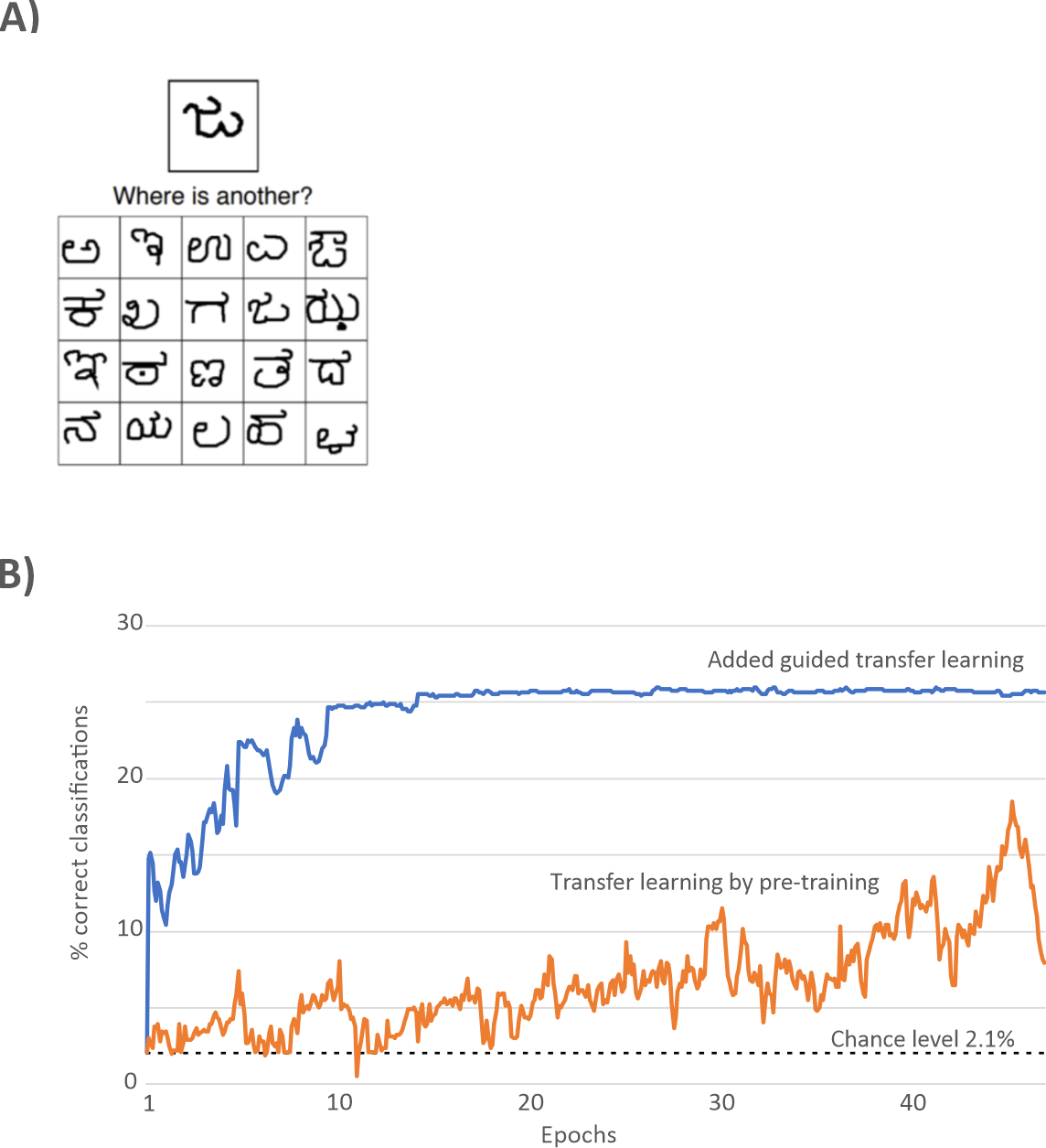}
\caption{A) Example task from the Omniglot data set: One example is given for training. Other examples of the same character need to be found. B) A representative example of the classification performance with the classical transfer learning by pre-training and extending it with guided transfer learning.}
\label{fig:long4}
\end{figure}

In further tests, we learned about the limitations of guided transfer learning. Guidance matrices quickly reached the maximum of the help they could provide. No matter how accurately we computed the guidance matrix, i.e. with how many examples were scouts trained and how many additional scouts were created, there was a limit on what the guidance matrix could do. The performance during guided transfer learning did not increase much beyond what is shown in Figure 4B (we always used single \(G\) matrices; it is currently unknown what would happen with more elaborate forms of guidance based on different \(G\) matrices applied at different epochs). On the other hand, we discovered an unexpected advantage of guided transfer learning: reducing the amount of training data and the number of scouts did not deteriorate the performance much. We did not need the full set of 10,000 examples to get the effect of guided transfer learning. We could get almost the same results with a lot fewer examples. Similarly, a reduction in the number of scouts trained did not significantly reduce the transfer learning performance either (results not shown). We concluded that: 1) our first attempt at guided transfer learning provides only limited advantages but 2) these advantages come at low costs in respect to the needed training data and computation. Guided transfer learning is not the ultimate solution for one-shot learning or learning in general. However, it can be quite helpful for avoiding overfitting in situations of insufficient data sizes.

\subsubsection{Avoiding local minima}
Certain tasks particularly suffer from local minima. An example is the aforementioned XOR function [8,9,10,11]. A small network can perform such tasks, but it is nearly impossible to find such a network by gradient descent. Meanwhile, bigger networks are hard to train. If one is ‘lucky’, the network stagnates at low performance for many epochs until it finally finds a solution, after which the increase in performance becomes much more rapid. This turning point can be referred to as the breaking point. However,  the networks reach no breaking point at all.

The question is then: Can we help the network to find breaking points easier? It is important to note that in these tasks the performance is not judged based on a testing data set. Instead, only the performance of the training data set is relevant. The problems are so hard in such tasks, and the local minimum is so debilitating, that even the training data set cannot be overfit. Therefore, we are not even asking the network to generalize; we only need to implement the mapping from the training data set. These types of difficulties are likely a key reason for the power law increase in the needed resources in real-life data [8].

\begin{figure}[H]
   \includegraphics[width=8.5cm]{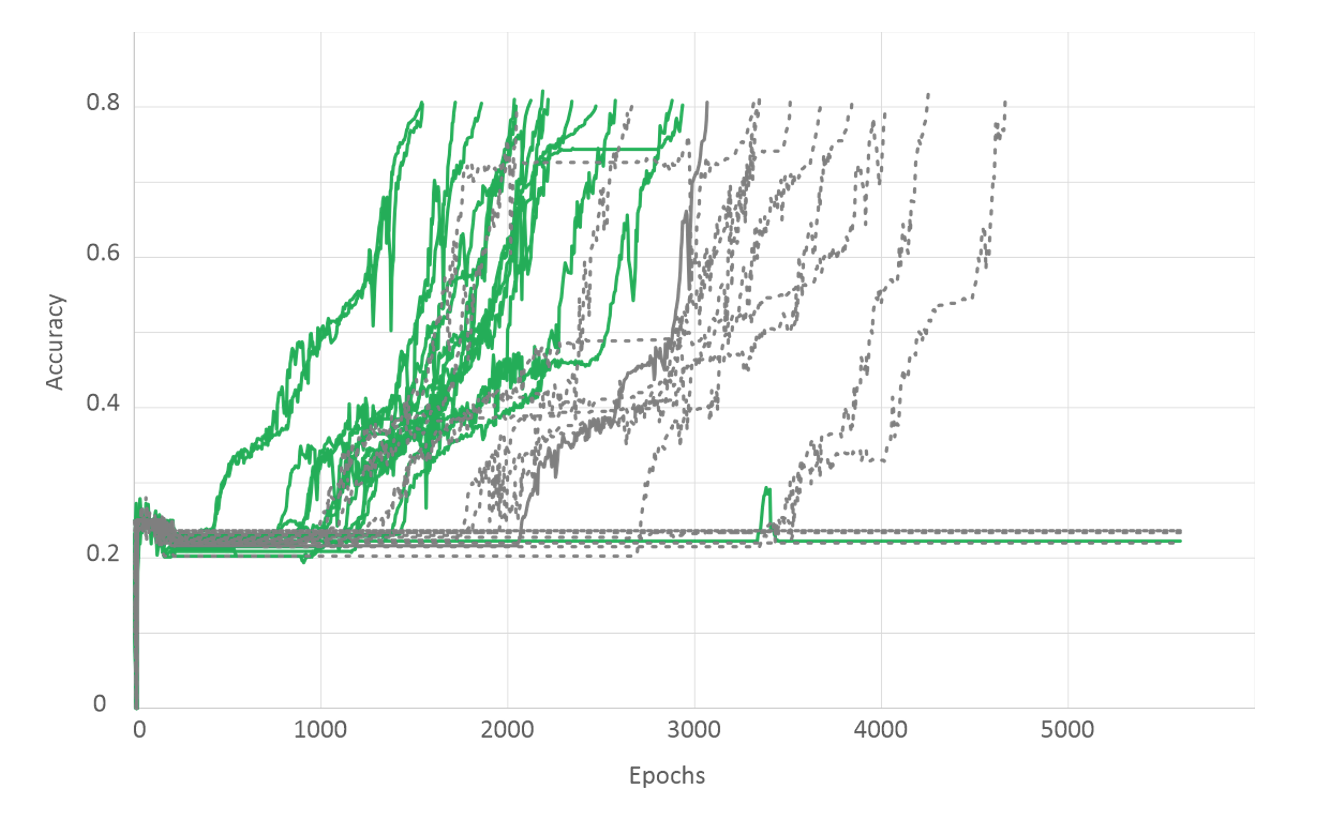}
   \caption{Example advantage in learning a difficult task with a guidance matrix. A difficult generalized XOR task was attempted to be learned 15 times using a guidance matrix (green, un-dashed lines) and 15 times without a guidance matrix (gray dashed lines). After quickly achieving about 20\% accuracy it became very difficult to find a solution to the task, requiring thousands of epochs. Application of a guidance matrix made the task easier.}
   \label{fig:long5}
\end{figure}

To apply guided transfer learning to these problems, we generated a problem with randomized input and output bits. This is a somewhat easier problem than pure generalized XOR mapping, but it is still a difficult task. 

To calculate a guidance matrix, we created scouts with a much smaller number of randomized mappings than the full task. We then applied \(G\) to new networks with the same type of random mapping task. Importantly, this problem did not include transfer learning by pre-training. Instead, the networks were randomly initialized each time a new task was present. Therefore, the networks started their guided transfer learning from a completely random state.

Our results showed that the guidance matrices were of great help. As seen in Figure 5, breaking through took much less time (fewer epochs) when guided transfer learning was applied as opposed to learning without this form of assistance. In fact, the training with guided transfer learning was so effective that no visible period of stagnation could be observed.

\subsubsection{Accumulation of knowledge}
Unexpectedly, we discovered one more interesting property of guided transfer learning: It allowed for the accumulation of knowledge. When artificial neural networks are trained on new data, they typically forget the information learned on old data. The weights change so much under the influence of new data that the knowledge based on previously learned data gets quickly lost. This is known as catastrophic forgetting [18,19,20] and is why all the training data have to be visited across the batches during each training epoch. Otherwise, if one would train for multiple epochs with one batch, and then multiple epochs with another batch, and so on, the performance would be much worse. Effectively, the network would learn only with the last batch, the contributions of the previous batches being mostly washed out during the more recent training epochs. 

Catastrophic forgetting is substantially reduced during guided transfer learning. To demonstrate this property, we performed a sequence of transfer learning procedures. We used the same setup as for one-shot learning above and repeatedly continued the training of a neural network with new single examples of the existing categories in each training. Each individual training lasted for a prespecified number of epochs, after which a new character example was randomly selected from the Omniglot data set. And so on. The testing was always made on the remaining 19 examples (there is a total of 20 examples per category in Omniglot data set). In other words, we made learning a bit more natural: In real lifes we do not see all the examples of say cars in one batch. Some examples of cars we have seen for the last time years ago. Nevertheless, our concept of cars remains to be influenced by these examples seen long ago. The question is whether GTL can enable artificial neural networks to do something similar.

Figure 6 shows a typical result. The classical transfer learning based solely on pre-training cannot benefit from such sequential learning from different examples. The performance does not increase with subsequent transfer learnings, or it increases only marginally. In contrast, when adding guided transfer learning on top of the algorithm, the performance exhibited a gradual stepwise increase with each new example added to the training. This was despite the training with one character per category at the time. This means that new training did not erase what the network learned before. Instead, the knowledge added up.

Again, there was a limit to how much this could be pushed; with repeated iterations, the returns diminished. For example, in the example in Figure 6 we did not obtain further improvements in the fifth iteration of one-shot transfer learning. Therefore, the network did forget eventually, but the forgetting took a lot longer. The network exhibited a high level of robustness on forgetting old knowledge.

\begin{figure}[H]
   \includegraphics[width=8cm]{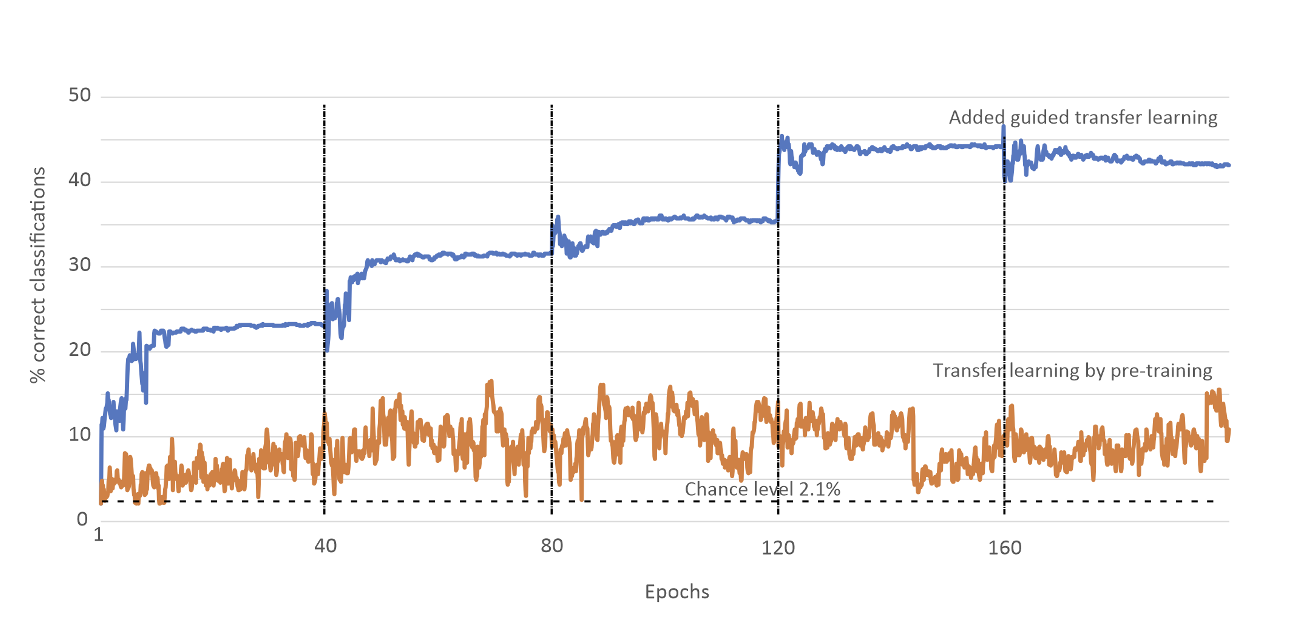}
   \caption{Repeated single-shot learning. At each block another example letter was picked for transfer training. Only by the addition of guided transfer learning was there a benefit from such sequential single-shot learning.}
   \label{fig:long6}
\end{figure}

This phenomenon can be explained in terms of constraints for learning. In multi-dimensional spaces, many good solutions exist for each example, but only a fraction of these solutions are good at the same time for other examples learned earlier. Moreover, these special solutions likely lie in the vicinity of where the model is when it begins with a new batch of learning. However, when training with one example only, not enough constraints are available to cause the model to stay in that vicinity; instead, as multiple epochs are repeated, the model wanders away. Typically, the missing constraints are added by considering more data at each epoch. Alternatively, we see that the guidance matrix can provide these constraints. The guidance matrix ensures that the model does not wander far away despite the multiple epochs. In terms of our mountain-lake metaphor from Figure 1B, learning without guidance matrices would be like following a long small stream of a watercourse that brings you far away from where you started from and as a result, you have a hard time finding a way back. The guidance matrix prevents you from walking far away.

\section{Discussion}

Guided transfer learning can be considered a method of learning how to learn. There are a number of other methods for learning how to learn. Some of them are focused on the problem of one-shot learning [21,22,23,24]; others on catastrophic forgetting [19,20]. There are also elaborate theories on how to generally address the problem of learning how to learn (e.g., [25]). What distinguishes guided transfer learning from these other methods is that it is a lightweight approach to this problem. Guided transfer learning is computationally cheap and can be applied to any deep learning network architecture. Also, as we have seen, it requires small amounts of data to learn how to learn. In contrast, many other methods require large computational resources and a lot of data. Even though these more elaborate methods can achieve better performance, they likely have poor scaling capabilities. In that way, their complexity often defies their purpose. This is likely why we do not find the elaborate methods being widely used in practice. The world seems to prefer simple, scalable methods. For example, transfer learning by pre-training is commonly applied, not because of its top performance but because of its simplicity and scalability. We think that Guided transfer learning should fall into this category.

These advantages are especially emphasized with models that have already squeezed out everything possible from their available resources. Large language models (e.g., GPT-3) cannot grow much larger because we are already near the limits of what the computer hardware and the available training data can offer. Further developments of such models may greatly benefit from Guided transfer learning as it is a scalable method for learning how to learn.
This is similar for models that process computer images. Their applications are often limited to peripheral uses, which are further limited in the computational resources available within, for example, autonomous driving vehicles. As one faces the pressure to pack as much intelligence into as small a model as possible, we think that guided transfer learning can be of great help.

It is important to note that guided transfer learning is not a panacea for all machine learning problems possible. Guided transfer learning does not solve everything and is not applicable in every situation. Often one will need to experiment with how to best use this technology, if at all. But of course, this same caveat applies to any other machine learning method. The no-free-lunch theorem in machine learning (e.g., [26]) states that no panacea solution can exist. Every method will have advantages and disadvantages. So too is the case with guided transfer learning.
It is also useful to keep in mind that a large enough training data set combined with a large enough model will likely beat guided transfer learning in terms of performance. Therefore, if data and model size are not an issue, one may be better off without guided transfer learning. Yet guided transfer learning may help even in these cases, if the problem is hard to break through. We saw this in the XOR example above.

A combination of a base network with a guiding matrix can be thought of as an intelligent “agent”. This agent already has knowledge about how to perform a certain task (stored in the base network) and the knowledge on how to adjust itself to a new related task (stored in the guidance matrices). These agents are light as they do not possess elaborate knowledge on how to learn or adjust themselves. Nevertheless, they can significantly extend our usage of deep learning models. Given that guided transfer learning applies to any neural network architecture and given the popularity of classical transfer learning, the creation of models with such advanced capabilities for transfer learning may be of great practical use for various AI developments. For this reason, we think guided transfer learning should apply to a wide range of problems.

\subsection{Acknowledgements}

We would like to thank Zilong Zhao for help with early implementations in TensorFlow and Katherine Munro for reading an earlier version of the manuscript. Conflicting interest disclosure: We have submitted a provisional patent at USPTO regarding a machine learning technology that is based on the presented work.

\subsection{References}

[1] LeCun, Yann, et al. "Gradient-based learning applied to document recognition." Proceedings of the IEEE 86.11 (1998): 2278-2324.

[2] Kaplan, Jared, et al. "Scaling laws for neural language models." arXiv preprint arXiv:2001.08361 (2020).

[3] Hoffmann, J., Borgeaud, S., Mensch, A., Buchatskaya, E., Cai, T., Rutherford, E., ... \& Sifre, L. (2022). Training Compute-Optimal Large Language Models. arXiv preprint arXiv:2203.15556.

[4] Thompson, N. C., Greenewald, K., Lee, K., \& Manso, G. F. (2020). The computational limits of deep learning. arXiv preprint arXiv:2007.05558.

[5] Bianco, Simone, et al. "Benchmark analysis of representative deep neural network architectures." IEEE Access 6 (2018): 64270-64277.

[6] Meir, Y., Sardi, S., Hodassman, S., Kisos, K., Ben-Noam, I., Goldental, A., \& Kanter, I. (2020). Power-law scaling to assist with key challenges in artificial intelligence. Scientific reports, 10(1), 1-7.

[7] Nikolić, D. (2022). Building Great Artificial Intelligence. The Handbook of Data Science and AI: Generate Value from Data with Machine Learning and Data Analytics, 239.

[8] Nikolić, D. (2023). Where is the mind within the brain? Transient selection of subnetworks by metabotropic receptors and G protein-gated ion channels. Computational Biology and Chemistry. arXiv preprint arXiv:2207.11249.

[9] Schmidhuber, J., \& Hochreiter, S. (1996). Guessing can outperform many long time lag algorithms.

[10] Linial, N., Mansour, Y., \& Nisan, N. (1993). Constant depth circuits, Fourier transform, and learnability. Journal of the ACM (JACM), 40(3), 607-620.

[11] Mansour, Y. (1994). Learning Boolean functions via the Fourier transform. In Theoretical advances in neural computation and learning (pp. 391-424). Springer, Boston, MA.

[12] Kingma, Diederik P., \& Jimmy Ba. "Adam: A method for stochastic optimization." arXiv preprint arXiv:1412.6980 (2014).

[13] Bozinovski, Stevo \& Fulgosi, Ante (1976). The influence of pattern similarity and transfer learning upon the training of a base perceptron B2. (original in Croatian) Proceedings of Symposium Informatica 3-121-5, Bled.

[14] Pratt, L. Y. (1993). Discriminability-based transfer between neural networks" (PDF). NIPS Conference: Advances in Neural Information Processing Systems 5. Morgan Kaufmann Publishers. pp. 204–211.

[15] Lake, B. M., Salakhutdinov, R., \& Tenenbaum, J. B. (2015). Human-level concept learning through probabilistic program induction. Science, 350(6266), 1332-1338.

[16] Lake, B. M., Salakhutdinov, R., \& Tenenbaum, J. B. (2019). The Omniglot challenge: a 3-year progress report. Current Opinion in Behavioral Sciences, 29, 97-104.

[17] Deng, L. (2012). The MNIST database of handwritten digit images for machine learning research [best of the web]. IEEE signal processing magazine, 29(6), 141-142.

[18] French, R. M. (1999). Catastrophic forgetting in connectionist networks. Trends in cognitive sciences, 3(4), 128-135.

[19] Kirkpatrick, J., Pascanu, R., Rabinowitz, N., Veness, J., Desjardins, G., Rusu, A. A., ... \& Hadsell, R. (2017). Overcoming catastrophic forgetting in neural networks. Proceedings of the national academy of sciences, 114(13), 3521-3526.

[20] Zeng, G., Chen, Y., Cui, B., \& Yu, S. (2019). Continual learning of context-dependent processing in neural networks. Nature Machine Intelligence, 1(8), 364-372.

[21] Vinyals, O., Blundell, C., Lillicrap, T., \& Wierstra, D. (2016). Matching networks for one shot learning. Advances in neural information processing systems, 29.

[22] Woodward, M., \& Finn, C. (2017). Active one-shot learning. arXiv preprint arXiv:1702.06559.

[23] Brown, T., Mann, B., Ryder, N., Subbiah, M., Kaplan, J. D., Dhariwal, P., ... \& Amodei, D. (2020). Language models are few-shot learners. Advances in neural information processing systems, 33, 1877-1901.

[24] Fei-Fei, L., Fergus, R., \& Perona, P. (2006). One-shot learning of object categories. IEEE transactions on pattern analysis and machine intelligence, 28(4), 594-611.

[25] Nikolić, D. (2015). Practopoiesis: Or how life fosters a mind. Journal of Theoretical Biology, 373, 40-61.

[26] Wolpert, D. H., \& Macready, W. G. (1997). No free lunch theorems for optimization. IEEE transactions on evolutionary computation, 1(1), 67-82.

\end{multicols}
\end{document}